\documentclass[conference]{IEEEtran}
\IEEEoverridecommandlockouts

\usepackage{cite}
\usepackage{amsmath,amssymb,amsfonts}
\usepackage{algorithmic}
\usepackage{textcomp}
\usepackage{graphicx}
\usepackage{tcolorbox}
\usepackage{listings}
\usepackage{xcolor}
\usepackage{tikz}
\usepackage{circuitikz}
\usepackage{url}
\usepackage{amssymb}
\usepackage{pifont}
\usepackage{multirow}       %
\usepackage{rotating}       %
\usepackage{makecell}       %
\usepackage{booktabs}       %
\usepackage[hidelinks]{hyperref}
\usepackage{cleveref}
\usepackage{hyperref}
\usepackage{xcolor}
\def\BibTeX{{\rm B\kern-.05em{\sc i\kern-.025em b}\kern-.08em
    T\kern-.1667em\lower.7ex\hbox{E}\kern-.125emX}}
\begin{document}

\title{Scalable Object Detection in the Car Interior \\
With Vision Foundation Models\\
}

\author{Sebastian Schmidt$^{1,2,\dagger}$, Bálint Mészáros$^{1,*,\dagger}$, Ahmet Firintepe$^{2,\dagger}$ and Stephan Günnemann$^{1}$%
\thanks{$^{1}$Technical University of Munich, School of Computation, Information and Technology, {\tt sebastian95.schmidt@tum.de }, {\tt bala.meszaros@gmail.com },  {\tt s.guennemann@tum.de}}%
\thanks{$^{2}$BMW Group 
        {\tt ahmet.firintepe@bmwgroup.com}}%
\thanks{$^{*}$Work has been conducted while employed at BMW}%
\thanks{$^{\dagger}$Equal Contribution}%
}

\newcommand{\cmark}{\textcolor{green}{\ding{51}}} %
\newcommand{\xmark}{\textcolor{red}{\ding{55}}}   %

\maketitle

\begin{abstract}
AI assistants operating in physical environments, such as vehicle interiors, must be capable of recognizing and localizing externally introduced and potentially unknown objects to ensure accurate and context-aware interaction. However, the computational limitations of on-board automotive systems restrict the deployment of such perception capabilities directly within the vehicle.
To address this limitation, we propose the novel Object Detection and Localization (ODAL) framework and present an empirical study of adapting and evaluating vision foundation models for in-cabin object detection and semantic localization. Our approach leverages a distributed architecture, splitting computational tasks between on-board and cloud. This design overcomes the resource constraints of running foundation models directly in the car. To benchmark model performance, we introduce ODALbench, a task-specific benchmark for comprehensive assessment of detection and localization.
Our analysis highlights potential for this approach in in-cabin perception, within the scope of our dataset. We compare the GPT-4o vision foundation model with the lightweight LLaVA 1.5 7B model and explore how task-tuning enhances the lightweight model performance. Remarkably, our task-tuned ODAL-LLaVA model achieves an ODAL$_{score}$ of 89\%, representing a 71 pp. improvement over its baseline performance and outperforming GPT-4o by nearly 20 pp. Furthermore, the task-tuned model maintains high detection accuracy while significantly reducing hallucinations, achieving an ODAL$_{SNR}$ three times higher than GPT-4o.
\end{abstract}

\section{INTRODUCTION}
The integration of artificial intelligence into the automotive industry has driven significant advances in applications such as driver assistance, autonomous navigation, and interior environment understanding. Among these, comprehending the car’s interior has become a key factor in improving user experience. Identifying externally introduced objects and their location inside the vehicle offers unique opportunities, enabling relevant recommendations, such as reminders about items when leaving the car. Traditional Convolutional Neural Network (CNN)-based classifiers excel at detecting objects they were trained on, but struggle with unseen classes and providing a general, human-readable description of their location. Given the wide range of possible user-introduced objects, in-cabin perception is inherently an open-vocabulary task, making a predefined label set nearly impossible.

This open-vocabulary setting motivates the use of vision–language foundation models, which offer strong generalization capabilities and can recognize a wider range of objects than conventional CNNs. In addition, due to their deep contextual understanding, they can semantically localize these objects within the interior. However, applying foundation models to open-vocabulary in-car perception entails important limitations: they require more computational power than smaller CNN- or Transformer-based classifiers, are difficult to fully deploy on vehicle hardware, and can produce free-form, sometimes hallucinated outputs that are not directly usable by downstream in-car systems.

\begin{figure}[t]
  \centering
  \includegraphics[width=0.47\textwidth]{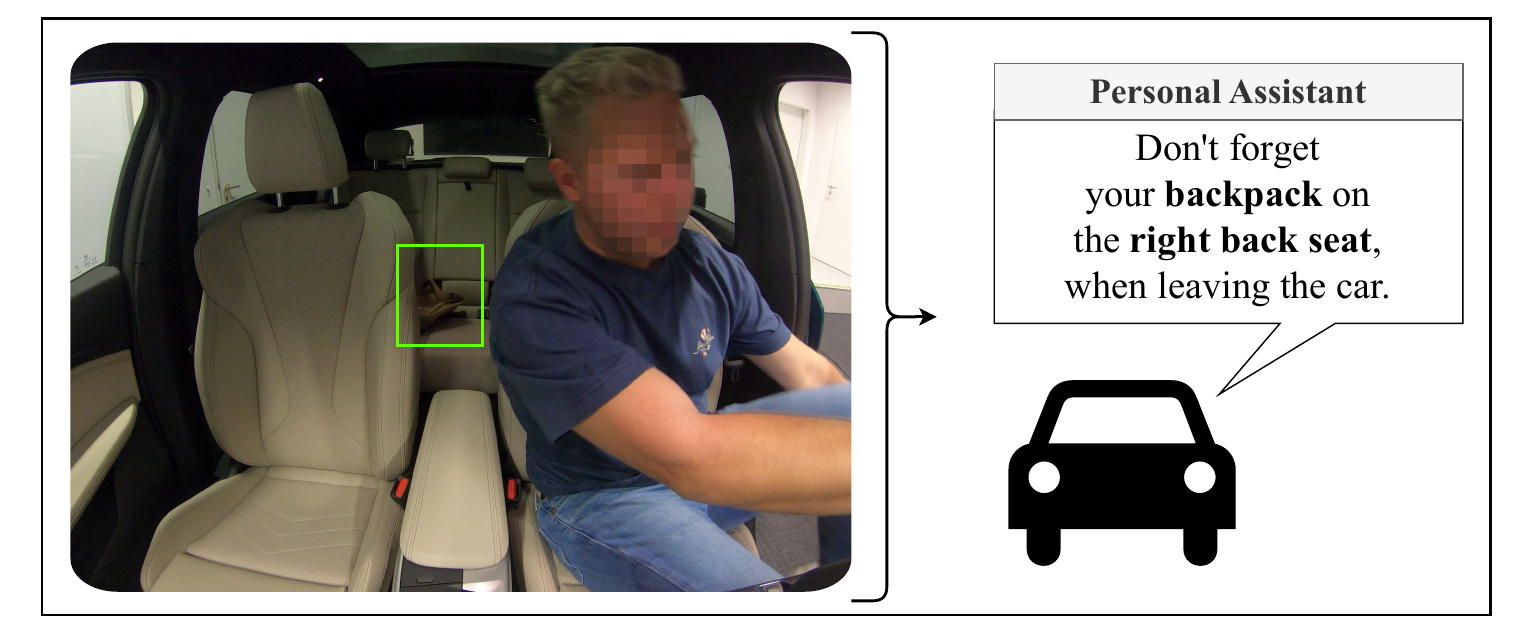}  %
  \vskip -0.4cm
  \caption{The vision of our framework, in which the car can understand the vehicle's interior. Based on the objects and their locations, smart recommendations can be provided to the driver and passengers.}
  \vskip -0.4cm  
  \label{fig:res}
\end{figure}

To address these challenges, we introduce the novel Object Detection and Localization (ODAL) framework as a reference pipeline and conduct an empirical evaluation of fine-tuning and prompting strategies for existing vision foundation models (VFMs) in this setting. By leveraging VFMs, the ODAL framework provides a robust solution for detecting and localizing objects within the car interior, enabling more accurate and reliable downstream task performance. Its architecture is distributed between the \textit{vehicle’s on-board system} and the \textit{cloud}, addressing the computational constraints of vehicle hardware while ensuring the privacy and security of raw images. In addition, we present a new benchmark, \textit{ODALbench}, for the proposed framework to provide a comprehensive evaluation of object detection and localization tasks.

Our work further highlights the framework’s potential to set new standards through a detailed analysis and comparison of state-of-the-art models. Specifically, we compare the performance of the GPT-4o VFM to a lightweight alternative, LLaVA 1.5 7B. While GPT-4o demonstrates strong baseline performance, we show that task-tuning the LLaVA 1.5 7B model for downstream tasks results in a threefold improvement on the ODAL$_{score}$ metric. Notably, the task-tuned model achieves superior accuracy while minimizing erroneous detections, thereby offering a more efficient and practical solution for real-world applications.

\begin{figure*}[t]
  \centering
  \vskip -0.1cm
  \includegraphics[width=0.83\textwidth]{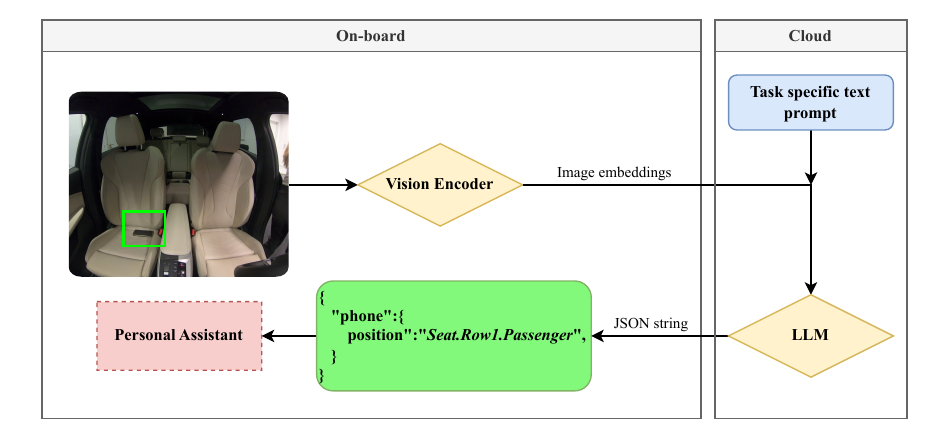}  %
  \vskip -0.5cm
  \caption{Overview of our ODAL framework. The system enables decoupled execution across on-board and cloud hardware: vision encoding runs in-car, only image embeddings are uploaded, and detection and localization are performed in the cloud. The resulting output can be used directly in the cloud or sent back to the car for other on-board applications, such as the Personal Assistant.}
  \label{fig:main}
  \vskip -0.35cm
\end{figure*}
Our contributions can be summarized as follows:
\begin{itemize}
    \item We introduce a car–cloud split framework that runs the image encoder on-board for privacy and low-latency feature extraction while offloading language-conditioned reasoning to the cloud.%
    \item We present a new, task-specific benchmark for interior scene understanding, focusing on human-understandable object detection and localization. %
    \item In an extensive evaluation, we provide performance improvement strategies enabling the LLaVA 1.5 7B model to outperform GPT-4o by the 2.9 fold SNR. 
\end{itemize}

\section{RELATED WORK}

\noindent\textbf{Traditional Object Detection and Localization:}
Classical object detectors \cite{tian2019fcos,zhou2019centernet} such as Faster R-CNN \cite{ren2015faster}, YOLO \cite{redmon2016you}, and DETR \cite{Carion2020DETR} achieve strong performance on predefined classes by training on large, domain-specific datasets. These methods operate in a closed-vocabulary regime: the label space is fixed at training time, preventing recognition of unseen categories that were not annotated in the dataset. Few can handle unknown classes~\cite{Schmidt2025}. Their outputs are restricted to bounding boxes and class IDs and therefore lack structured, human-readable location descriptions (e.g., ``on the front passenger seat''), which are essential for in-car assistant systems. As a result, classical detectors are ill-suited for dynamic automotive interiors where user-introduced objects are unknown a priori, and where semantic location descriptors must integrate with downstream automotive HMI components.
\vspace{2pt}

\noindent\textbf{Vision Foundation Models:}
The limitations of conventional detectors have led to the rise of vision foundation models (VFMs) \cite{radford2021learning,alayrac2022flamingo,dai2023instructblip,liu2023visual,liu2024improved}, which combine visual and linguistic understanding through large-scale image--text pretraining. Models such as CLIP \cite{radford2021learning} align image and text embeddings in a shared latent space, enabling zero-shot recognition of unseen objects and generalization beyond fixed label sets, making them suitable for diverse and unpredictable in-cabin environments.

Recent multimodal systems, including Flamingo \cite{alayrac2022flamingo}, InstructBLIP \cite{dai2023instructblip}, LLaVA \cite{liu2023visual,liu2024improved}, and GPT-4o extend VFMs with instruction-following capabilities and spatial reasoning. These models can interpret object identity, relationships, and locations and produce both natural-language and structured outputs, supporting tasks such as cabin inventory, semantic scene description, and forgotten-item alerts. Their adaptability via prompting, rather than retraining, allows flexible integration into evolving automotive HMI ecosystems.
\vspace{2pt}

\noindent\textbf{Open-Vocabulary Perception:}
The large-scale training of VFMs enables open-vocabulary perception, allowing recognition and reasoning beyond fixed label sets. Vision--language encoders such as CLIP serve as key components in open-vocabulary segmentation and detection frameworks \cite{Cho2024CATSeg}, while Visual Question Answering (VQA) models \cite{Antol2015VQA} demonstrate similar multimodal grounding capabilities.
\vspace{2pt}

Our work leverages these open-vocabulary properties to provide structured perception within the vehicle cabin. By grounding visual features in language, the proposed ODAL framework identifies arbitrary interior objects and produces machine-readable JSON descriptions for downstream in-car processing.

\section{OBJECT DETECTION AND LOCALIZATION FRAMEWORK}
To address the open-vocabulary \cite{huang2024renovatingnames, miller2025openset} challenges of unknown objects in the interior, the need for human-understandable and machine-usable outputs, and the computational limits of in-vehicle hardware, we propose the ODAL framework.
ODAL provides a direct, human-understandable output that relates objects in the interior scene. 
We place the visual encoder on board to avoid raw-image transfer, reduce latency to features, and ensure that only compact image embeddings are transmitted. Cloud-side LLM/VLM modules perform semantic detection and localization conditioned on the embeddings and prompts.
\vspace{2pt}

\noindent\textbf{Automotive Challenges:}

While VFMs offer strong generalization, their deployment in embedded automotive systems remains constrained by computational and reliability factors. Unlike lightweight CNNs, VFMs rely on large transformer backbones \cite{dehghani2023scaling}, requiring distributed computation that keeps privacy-sensitive tasks local while offloading semantic reasoning to the cloud. 

Automotive environments impose additional demands: high-resolution imagery for small-object detection (e.g., keys, wallets), accurate spatial localization in confined spaces, and consistent structured output for downstream processing.
ODAL must therefore satisfy:
(1) \textbf{High recognition and semantic localization accuracy}, objects should be recognized and localized in the interior semantics;
(2) \textbf{Low hallucination rate}, avoiding false detections for in-car interaction contexts; and
(3) \textbf{Strict instruction adherence and schema compliance}, near 100\% conformity to predefined JSON formats, ensuring consistent and interpretable outputs.

VFMs inherently support these requirements through generalization, contextual reasoning, and language-conditioned adaptability. ODAL builds on these strengths to enable scalable, privacy-aware interior perception.

\vspace{2pt}
\noindent\textbf{Framework:}
The ODAL framework extends LLaVA~1.5 \cite{liu2024improved}, integrating a pre-trained language model with a visual encoder to enable multimodal understanding of automotive interiors. The framework builds upon the LLaVA structure through task-specific configurations and parameter-efficient task-tuning using Low-Rank Adaptation (LoRA) \cite{hu2021lora}. These enhancements address the core requirements outlined above: \textit{high accuracy}, \textit{low hallucination rate}, and \textit{strict schema compliance}.

To adapt the generic LLaVA~1.5~7B model to the ODAL task, we employ task-tuning within the Hugging Face ecosystem, leveraging \texttt{Transformers} \cite{wolf2020transformers}, \texttt{TRL} \cite{vonwerra2022trl}, and \texttt{PEFT} \cite{peft}. LoRA adapters inject task-specific knowledge while keeping the number of trainable parameters small. Two complementary tuning strategies are explored:

\begin{itemize}
\item \textbf{Selective Task-Tuning:} Updates only the attention projection layers, significantly reducing computation while adapting the model to ODAL’s JSON-based detection and localization task.
\item \textbf{Comprehensive Task-Tuning:} Extends updates to all linear layers via LoRA, achieving higher accuracy at increased training cost.
\end{itemize}

These strategies balance performance and efficiency: selective tuning provides a lightweight baseline, whereas comprehensive tuning maximizes task accuracy when additional computational resources are available. Together, they ensure consistent, schema-compliant outputs and minimal hallucination across diverse in-vehicle conditions.

For efficient inference, ODAL employs the LLaVA~1.5 backbone, combining the Vicuna~7B~1.5 language decoder \cite{vicuna2023} with the CLIP ViT-L/14 encoder \cite{radford2021learning} through a two-layer MLP connector. CLIP encodes 336×336 images into compact embeddings aligned with the language model for unified visual–textual reasoning. To further reduce memory and latency, 4-bit quantization using \texttt{BitsAndBytes} is applied during training and inference, supporting resource-constrained inference on embedded systems.

The framework accepts multimodal prompts (image + text) and outputs structured JSON responses enumerating detected objects, spatial positions, and semantic attributes. This format enables direct downstream integration, for instance by passing detections to the \textit{BoardNet} subsystem for decision-making, without additional text parsing.

Finally, ODAL’s modular design supports incremental adaptation to new perception tasks and scaling across model configurations (7B, 13B, and 34B) or alternative language models such as Vicuna \cite{vicuna2023}, and LLaMA \cite{touvron2023llama}. The framework is designed to operate in hybrid mode (on-board encoding with cloud inference) or fully on-board, though this paper reports offline experiments and does not include real-vehicle deployment validation. A full overview of our framework can be seen in Fig. \ref{fig:main}.
\vspace{2pt}

\noindent\textbf{Dataset:}
We collected and labeled a dataset to evaluate and task-tune the models. It contains 223 images of a real vehicle interior with thirteen everyday personal object categories (e.g., backpacks, laptops, wallets) placed at predefined cabin locations, collected inside a BMW using a GoPro Hero7 Black at \(4000 \times 3000\) pixels, ensuring variability through different object orientations and a natural distribution of object placements, meaning the more frequent positions were more frequently present in the dataset. For each image, an annotator varied which objects are present and where they are positioned to mimic in-car scenarios. Annotations follow the ODAL JSON schema, specifying for every object its semantic location and visibility; an example is shown in Fig.~\ref{fig:json}. In addition, objects with a low frequency have been up-sampled. The data was then split into 80\% training set and 20\% validation set.

\begin{figure}[h]
\vskip -0.2cm
    \centering
    \begin{tcolorbox}[colframe=black,
        colback=white,
        coltitle=black, 
        sharp corners, 
        boxrule=0.7pt,
        top=-4pt,
        bottom=-4pt,
        ]
        \begin{lstlisting}[language=Python, 
            basicstyle=\ttfamily, 
            keywordstyle=\color{blue}, 
            stringstyle=\color{red}, 
            commentstyle=\color{gray}, 
            numbers=left, 
            numberstyle=\tiny\color{gray}, 
            stepnumber=1, 
            numbersep=5pt, 
            breaklines=true]
{
  "backpack": {
    "position": "Seat.Row2.Middle"
  }
}
        \end{lstlisting}
    \end{tcolorbox}
    \vskip -0.4cm
    \caption{Example of the ODAL JSON schema for the $backpack$ object.}
    \label{fig:json}
      \vskip -0.2cm
\end{figure}

To address the limited amount of data available for task-tuning, three levels of data augmentation were applied: (1) \textit{no augmentation}, (2) \textit{basic augmentation}, which included \textit{rotation}, \textit{flipping}, and \textit{brightness adjustment}, and (3) \textit{extensive augmentation}, which, in addition to the basic techniques, incorporated affine transformations and sharpness adjustments.

\section{EVALUATION}
\begin{figure*}[thpb]
 \vskip -0.2cm  
  \centering
    \includegraphics[scale=0.96]{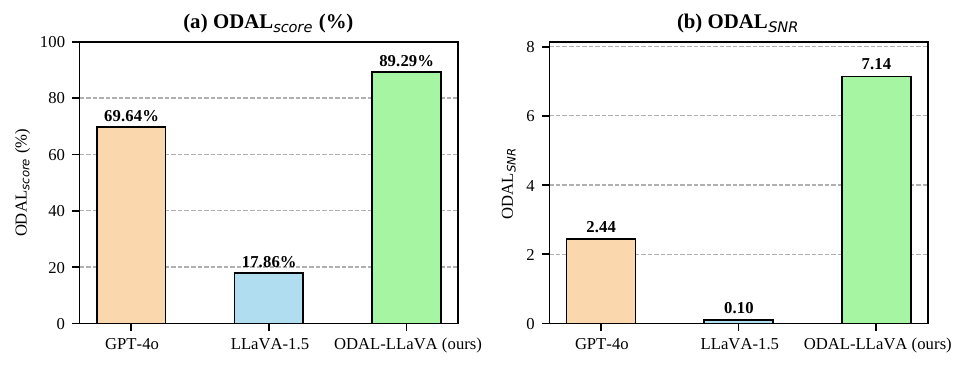}
    \vskip -0.4cm  
  \caption{Comparison of GPT-4o, LLava-1.5, and our ODAL-LLaVA on two different metrics. a) Plot of model performance for ODAL$_{score}$. b) Plot of model performance for ODAL$_{SNR}$.}
  \label{fig:comarasion}
    \vskip -0.4cm  
\end{figure*}

\textbf{Evaluation Pipeline:} Evaluating object detection and localization capabilities in generative vision–language models is challenging because LLMs often produce semantically equivalent yet syntactically diverse outputs. This variability limits the utility of classical precision–recall and text-similarity metrics. To address this, we introduce ODALbench, a two-stage evaluation framework based on an "LLM-as-a-judge" paradigm.

\textit{Step 1: Structured Prediction Extraction.}
Each model-generated response is passed to a dedicated judge LLM—GPT-4o in our experiments, which extracts canonicalized information including (i) detected objects, (ii) hallucinated objects, and (iii) localization correctness at predefined interior slots. Gold labels were human-verified, and the judge operated only on canonicalized predictions. The model-under-test (LLaVA-1.5, ODAL-LLaVA, or GPT-4o) is fully isolated from the judge. In this constrained extraction setting, judge-induced errors are negligible relative to model variance.

\textit{Step 2: Metric Computation.}
The extracted structured predictions are processed by an evaluation script to compute ODALbench metrics. This separation ensures deterministic scoring despite generative variability.

\noindent\textbf{ODALbench:}
Classic metrics such as precision, recall, and mAP assume set-based label matching but cannot handle text-based semantic localizations. Our target is a structured JSON with localizations at discrete interior positions.
In addition, these metrics do not capture additional properties of generative models, like hallucination.
Existing methods struggle to capture the structure of textual outputs, where positioning is simply correct or incorrect, rather than based on similarity.
To provide a more detailed evaluation of model performance in detection and localization tasks, while also accounting for hallucination, we introduce a novel, task-specific benchmark.

Our introduced ODALbench consists of two metrics: ODAL$_{score}$ and ODAL$_{SNR}$. The ODAL$_{score}$ quantifies the detection and localization performance for each frame. While the ODAL$_{SNR}$ (Signal-to-Noise Ratio) focuses on hallucinations. 

The ODAL$_{score}$ quantifies frame-level performance by combining detection and localization correctness, while penalizing hallucinated objects. Each correctly detected and localized object receives one point; correct detection with incorrect localization receives half a point. Frames with no correct detections receive a default score of one. A penalty of -1 is applied for each hallucinated object. The score for frame $f$ is:
\[
S_f = \sum_{i=1}^{N} \left( d_i \cdot l_i + d_i \cdot \frac{(1 - l_i)}{2} \right) + \delta_{0, D}
\]
Where \(d_i\) indicates correct detection (1 for yes, 0 for no), \(l_i\) indicates correct localization (1 for yes, 0 for no), and \(\delta_{0,D}\) is a special case that awards one point if no object is detected correctly.

To also evaluate model hallucination, we introduce a Signal-to-Noise Ratio metric, ODAL$_{SNR}$.
This metric quantifies the ratio between the number of correct detections ($C$) and hallucinations ($H$) for a given dataset:
\[
ODAL_{SNR} = \frac{C}{H}
\]
A higher ODAL$_{SNR}$ signals better performance, but the metric may converge to infinity with no hallucinations, which can occur if there is division by zero. This is handled by taking the maximum possible value of ODAL$_{SNR}$. For models with hallucinations, ODAL$_{SNR}$ falls within the range of zero to the maximum possible number of correct detections in the dataset.

ODALbench offers a comprehensive and interpretable evaluation of a model’s detection and localization capabilities, highlighting both accuracy and the quality of detections.
The earlier reference to “semantically equivalent entities” refers to name canonicalization in the judge step; after normalization, ODALbench evaluates binary detection and localization by design. The omission of a similarity term in ${S_f}$ is intentional.
\vspace{2pt}

\noindent\textbf{Experiment Setup:}
Throughout this study, several experimental setups were explored using our recorded dataset. These included the previously mentioned selective and comprehensive task-tuning approaches, which targeted different aspects of the model's parameters. Unless stated otherwise, tuning used LoRA adapters on the specified modules while keeping the remaining backbone weights frozen. Additionally, two distinct prompt configurations were evaluated: the model's user prompt was customized as follows, while the system prompt remained unchanged. All models were evaluated on the same prompts and ODAL JSON schema, and were scored by the same ODALbench pipeline:

\begin{itemize}
    \item \textbf{Detailed prompt} (V1): Provides comprehensive task instructions, including a description of the environment where objects are to be detected and localized (a car interior), naming conventions for possible object positions, and the expected output JSON string format with examples.  
    \item \textbf{Minimalistic prompt} (V2): Briefly defines the environment (a car interior) and the task of detecting and localizing objects, while omitting naming conventions and the expected response structure with examples.  
\end{itemize}
\vspace{2pt}

\noindent\textbf{Results:}
The state-of-the-art \textbf{GPT-4o} model, the \textbf{baseline LLaVA 1.5} model, and our task-tuned \textbf{ODAL-LLaVA} variants were evaluated based on three aspects: ODAL$_{score}$, ODAL$_{SNR}$, and the accuracy with which the models reproduced the expected JSON string format. 
The models were task-tuned using various configurations to assess how changes in the trained parameters or input prompts affected performance. 
In addition to the previously discussed selective and comprehensive task-tuning approaches, models were trained with different setups, including task-tuning only the LLM component or both the LLM and Vision Encoder (VE). 
The initial iterations of model task-tuning focused on the different prompt types used. 
The first version employed a detailed prompt, while the second used a minimalistic prompt. 
In the first iteration, all three augmentation levels were compared. 
In the second iteration, however, only the basic augmentation level was used, as it demonstrated the best performance in the first iteration, as shown in Table \ref{tab:augment}.

\begin{table}[tbh]
\vskip -0.2cm
\centering
\caption{Results of the trained ODAL-LLaVA model variants with different augmentation strategies. All models use the same task-tuning settings: Vision Encoder and LLM are task-tuned in a selective fashion while using a detailed prompt.}\label{tab:augment}
\vskip -0.1cm
\setlength{\tabcolsep}{5.2pt}
\begin{tabular}{c|c|ccc}
\toprule
\multirow{3}{*}{\rotatebox[origin=c]{90}{\makecell{Version}}} & 
\multirow{3}{*}{\rotatebox[origin=c]{90}{\makecell{Augment. \\ Type}}} & 
\multicolumn{3}{c}{ODALbench} \\
\addlinespace[14pt]
& & {\scriptsize ODAL$_{score} (\%)$} & {\scriptsize ODAL$_{SNR}$} & {\scriptsize JSON Rate (\%)} \\
\midrule
V1 & Basic     & \textbf{74.10}  & \textbf{1.2205}  & \textbf{97.72} \\
V1 & No        & 72.32  & 1.0125  & 61.36 \\
V1 & Extensive & 19.64  & 0.55    & 6.81 \\
\bottomrule
\end{tabular}
\vskip -0.4cm
\end{table}

Using this augmentation variant, we evaluate the task-tuning strategy in Table \ref{tab:odal_models}. The results show the strong performance of ODAL-LLaVA when task-tuning both the model parts of VE and LLM on the data with basic augmentation by using a minimalistic prompt.

The evaluation of the models began by assessing their ability to generate responses in the expected format. The GPT-4o model achieved perfect performance, with a 100\% success rate in producing responses that conformed to the required format. In contrast, the baseline LLaVA 1.5 model struggled significantly, frequently generating overly detailed and verbose responses that could not be parsed as valid JSON strings. However, after task-tuning, the LLaVA 1.5 model demonstrated a remarkable transformation, achieving 100\% success in generating the correct response format, effectively matching the performance of GPT-4o.

We then compared the models based on their ODAL$_{score}$ percentage, which reflects the accuracy of object detection and localization. As shown in Fig. \ref{fig:comarasion}, GPT-4o achieved 69.64\% of the maximum possible ODAL$_{score}$, a strong result in terms of object localization. 
In contrast, the baseline LLaVA 1.5 model performed poorly, attaining only 17.86\% of the maximum score. 
Task-tuning the LLaVA 1.5 model resulted in a dramatic improvement, increasing its ODAL$_{score}$ percentage to 89.29\%. 
This performance not only represents a substantial gain over the baseline but also surpasses the GPT-4o model by nearly 20\%. 
These results clearly demonstrate that our task-tuning process significantly enhanced the model's ability to understand and localize objects within the car interior, marking a clear improvement over existing models.

\begin{table}[tbh]
\vskip -0.2cm
\centering
\caption{Results of the trained ODAL-LLaVA model variants with basic augmentation. The LLM model component was trained in all cases. (V1: detailed prompt, V2: minimalistic prompt)}\label{tab:odal_models}
\vskip -0.1cm
\setlength{\tabcolsep}{5.2pt}
\begin{tabular}{c|cc|ccc}
\toprule
\multirow{4}{*}{\rotatebox[origin=c]{90}{\makecell{Version}}} & 
\multicolumn{2}{c|}{Task-Tuning} & \multicolumn{3}{c}{ODALbench} \\
\addlinespace[3pt]
& \parbox[t]{2mm}{\multirow{2}{*}{\rotatebox[origin=c]{90}{\makecell{Vision\\Encoder}}}} & 
\multirow{2}{*}{\rotatebox[origin=c]{90}{\makecell{Compre-\\hensive}}} & 
\multicolumn{3}{c}{} \\
\addlinespace[14pt] %
& & & {\scriptsize ODAL$_{score} (\%)$} & {\scriptsize ODAL$_{SNR}$} & {\scriptsize JSON Rate (\%)} \\
\midrule
V2 & \cmark & \cmark & \textbf{76.79} & \textbf{7.1428} & \textbf{100} \\
V1 & \cmark & \cmark & 68.75 & 5.2777 & 95.45 \\
V1 & \xmark & \cmark & 69.64 & 5.0000 & 93.18 \\
V2 & \xmark & \cmark & 70.54 & 3.8214 & \textbf{100} \\
V2 & \xmark & \xmark & 41.07 & 1.8518 & 31.82 \\
V2 & \cmark & \xmark & 33.04 & 1.5441 & 47.73 \\
V1 & \xmark & \xmark & 21.43 & 1.3870 & 97.72 \\
V1 & \cmark & \xmark & 13.39 & 1.2205 & 97.72 \\
\bottomrule
\end{tabular}
\vskip -0.4cm
\end{table}

The final metric used for evaluation was the ODAL$_{SNR}$, which quantifies the frequency of hallucinated objects relative to correct detections. 
Ideally, the ODAL$_{SNR}$ should approach the number of correctly detectable objects, which in this case would be 56. 
As shown in Fig. \ref{fig:comarasion}, GPT-4o achieved an ODAL$_{SNR}$ of 2.44, indicating that for every 2.5 correct detections, there was one hallucination. The baseline LLaVA 1.5 model performed poorly, with an ODAL$_{SNR}$ of just 0.1, meaning it hallucinated nearly 10 times as many objects as it correctly detected. In contrast, after task-tuning, the LLaVA 1.5 model demonstrated a substantial improvement, achieving an ODAL$_{SNR}$ of 7.14. This represents a more than threefold increase over GPT-4o, highlighting a dramatic reduction in the frequency of hallucinations and a clear improvement in model quality.
Although the theoretical upper bound equals the maximum correct detections in the set, this value already reflects substantially reduced false positives in practice.

\textit{Overall}, the results underscore the substantial improvements achieved through task-tuning, which not only brought the performance of LLaVA 1.5 in line with GPT-4o but actually surpassed it in key areas such as object localization and hallucination reduction. These improvements showcase the efficacy of our custom task-tuning approach in improving model performance on our single-vehicle dataset, while broader generalization remains future work.

\section{LIMITATIONS}
While ODAL demonstrates promising results, several limitations remain. 
\textbf{(i) Decoupled execution.} Splitting vision encoding on-board and language-driven reasoning in the cloud introduces dependence on communication reliability, latency, and bandwidth. Temporary outages can yield stale or deferred results despite batching and last-result fallback, and strict end-to-end real-time guarantees are not provided. However, this can be mitigated by a full on-board calculation, increasing computational demands on the vehicle.
\textbf{(ii) Dataset scope.} The evaluation uses a small-scale dataset (223 images, 13 objects) recorded in a single vehicle configuration with five predefined interior locations. Upsampling mitigates imbalance but may introduce a distribution shift. Generalization to other cabins, cameras, lighting, or object sets is not established. 
\textbf{(iii) Metric design.} ODALbench targets structured outputs with binary localization correctness; it intentionally omits graded similarity and is therefore not directly comparable to box-based precision–recall. $ODAL_{SNR}$ becomes ill-defined at $H{=}0$ (handled by capping), which complicates comparisons near zero hallucinations. 
\textbf{(iv) LLM-as-judge.} The judge step relies on GPT-4o for normalization and extraction; although we observed negligible judge-induced errors, residual extraction mistakes and version drift are possible and may affect scores, and the judge may introduce systematic bias toward certain wording or error patterns despite canonicalization.
\textbf{(v) System constraints.} On-board encoding still consumes compute and energy, and the feasible embedding rate is bounded by hardware and link capacity. Privacy is improved by transmitting embeddings instead of raw images, but a formal leakage analysis of embeddings is out of scope.

\section{CONCLUSION}
In this paper, we introduced ODAL, a context-aware in-car object detection and localization framework built upon the LLaVA-1.5 7B vision foundation model. 
Leveraging a distributed architecture, the framework enables computationally efficient execution across both vehicle hardware and cloud environments, addressing the challenges of limited on-board processing power while ensuring privacy compliance.  
 
To this end, we proposed a new benchmark, ODALbench, to facilitate the evaluation.
This benchmark introduces two key metrics: ODAL$_{score}$, which quantifies detection and localization accuracy, and ODAL$_{SNR}$, a signal-to-noise ratio that evaluates the impact of hallucinations on model response quality. 
These metrics provide a comprehensive assessment of the framework’s performance in real-world scenarios.  

Furthermore, we evaluated the baseline and task-tuned versions of the framework and compared their performance to the state-of-the-art GPT-4o model. 
Our results indicate that, although the baseline LLaVA-1.5 model significantly underperforms GPT-4o, task-tuning substantially improves its performance. 
The task-tuned ODAL-LLaVA model achieved an ODAL$_{score}$ of 89\%, representing a 71\% absolute improvement over its baseline performance and surpassing GPT-4o by nearly 20\%. 
Additionally, it achieved a threefold improvement in ODAL$_{SNR}$, indicating a substantial reduction in hallucinations. 
These findings underline the effectiveness of adapting pre-trained models for domain-specific tasks, particularly in automotive applications, where, due to limited computational resources, models need to be distributed between the vehicle's on-board system and the cloud.

While this study demonstrates strong performance, in future work, we aim for in-car deployment and extend the validation to more scenarios, with additional objects and more challenging scenes. 
Further, we plan to explore the impact of domain-specific task-tuning to the model’s general knowledge.

\addtolength{\textheight}{-12cm}   %


\begin{thebibliography}{99}

\bibitem{ren2015faster}
S. Ren, K. He, R. Girshick, and J. Sun, ``Faster R-CNN: Towards Real-Time Object Detection with Region Proposal Networks,'' in \textit{Proc. Adv. Neural Inf. Process. Syst. (NeurIPS)}, 2015.

\bibitem{redmon2016you}
J. Redmon, S. Divvala, R. Girshick, and A. Farhadi, ``You Only Look Once: Unified, Real-Time Object Detection,'' in \textit{Proc. IEEE/CVF Conf. Comput. Vis. Pattern Recognit. (CVPR)}, 2016.

\bibitem{tian2019fcos}
Z. Tian, C. Shen, H. Chen, and T. He, ``FCOS: Fully Convolutional One-Stage Object Detection,'' in \textit{Proc. IEEE/CVF Int. Conf. Comput. Vis. (ICCV)}, 2019.

\bibitem{zhao2019object}
Z. Zhao, P. Zheng, S. Xu, and X. Wu, ``Object Detection with Deep Learning: A Review,'' \textit{IEEE Trans. Neural Netw. Learn. Syst.}, 2019.

\bibitem{radford2021learning}
A. Radford \textit{et al.}, ``Learning Transferable Visual Models From Natural Language Supervision,'' in \textit{Proc. Int. Conf. Mach. Learn. (ICML)}, 2021.

\bibitem{bommasani2021opportunities}
R. Bommasani \textit{et al.}, ``On the Opportunities and Risks of Foundation Models,'' \textit{arXiv preprint arXiv:2108.07258}, 2021.

\bibitem{alayrac2022flamingo}
J.-B. Alayrac \textit{et al.}, ``Flamingo: a Visual Language Model for Few-Shot Learning,'' in \textit{Proc. Adv. Neural Inf. Process. Syst. (NeurIPS)}, 2022.

\bibitem{dai2023instructblip}
W. Dai \textit{et al.}, ``InstructBLIP: Towards General-Purpose Vision-Language Models with Instruction Tuning,'' in \textit{Proc. Adv. Neural Inf. Process. Syst. (NeurIPS)}, 2023.

\bibitem{zhou2019centernet}
K. Duan \textit{et al.}, ``CenterNet: Keypoint Triplets for Object Detection,'' in \textit{Proc. IEEE/CVF Int. Conf. Comput. Vis. (ICCV)}, 2019.

\bibitem{liu2023visual}
H. Liu, C. Li, Q. Wu, and Y. J. Lee, ``Visual Instruction Tuning,'' in \textit{Proc. Adv. Neural Inf. Process. Syst. (NeurIPS)}, 2023.

\bibitem{liu2024improved}
H. Liu, C. Li, Y. Li, and Y. J. Lee, ``Improved Baselines with Visual Instruction Tuning,'' in \textit{Proc. IEEE/CVF Conf. Comput. Vis. Pattern Recognit. (CVPR)}, 2024.

\bibitem{hu2021lora}
E. J. Hu \textit{et al.}, ``LoRA: Low-Rank Adaptation of Large Language Models,'' \textit{arXiv preprint arXiv:2106.09685}, 2021.

\bibitem{dehghani2023scaling}
M. Dehghani \textit{et al.}, ``Scaling Vision Transformers,'' \textit{arXiv preprint arXiv:2302.05442}, 2023.


\bibitem{touvron2023llama}
H. Touvron \textit{et al.}, ``LLaMA: Open and Efficient Foundation Language Models,'' \textit{arXiv preprint arXiv:2302.13971}, 2023.

\bibitem{vicuna2023}
W.-L. Chiang, Z. Li, Z. Lin, Y. Sheng, Z. Wu, H. Zhang, L. Zheng, S. Zhuang, Y. Zhuang, J. E. Gonzalez, I. Stoica, and E. P. Xing, ``Vicuna: An Open-Source Chatbot Impressing GPT-4 with 90\%* ChatGPT Quality,'' Mar. 2023. [Online]. Available: \url{https://lmsys.org/blog/2023-03-30-vicuna/}

\bibitem{miller2025openset}
D. Miller \textit{et al.}, ``Open-Set Recognition in the Age of Vision-Language Models,'' in \textit{Proc. Eur. Conf. Comput. Vis. (ECCV)}, 2024.

\bibitem{huang2024renovatingnames}
H. Huang \textit{et al.}, ``Renovating Names in Open-Vocabulary Segmentation Benchmarks,'' \textit{arXiv preprint arXiv:2403.09593}, 2024.

\bibitem{wolf2020transformers}
T. Wolf \textit{et al.}, ``Transformers: State-of-the-Art Natural Language Processing,'' in \textit{Proc. Conf. Empirical Methods Natural Lang. Process.: Syst. Demonstrations (EMNLP)}, 2020.

\bibitem{vonwerra2022trl}
L. von Werra \textit{et al.}, ``TRL: Transformer Reinforcement Learning,'' GitHub repository, 2020. [Online]. Available: \url{https://github.com/huggingface/trl}.

\bibitem{peft}
S. Mangrulkar \textit{et al.}, ``PEFT: State-of-the-art Parameter-Efficient Fine-Tuning methods,'' 2022. [Online]. Available: \url{https://github.com/huggingface/peft}.

\bibitem{Antol2015VQA}
S. Antol, A. Agrawal, J. Lu, M. Mitchell, D. Batra, C. L. Zitnick, and D. Parikh,
``VQA: Visual Question Answering,''
in \textit{Proc. IEEE Int. Conf. Comput. Vis. (ICCV)}, 2015.

\bibitem{Cho2024CATSeg}
S. Cho, H. Shin, S. Hong, A. Arnab, P. H. Seo, and S. Kim,
``CAT-Seg: Cost Aggregation for Open-Vocabulary Semantic Segmentation,''
in \textit{Proc. IEEE/CVF Conf. Comput. Vis. Pattern Recognit. (CVPR)}, 2024.

\bibitem{Carion2020DETR}
N.~Carion, F.~Massa, G.~Synnaeve, N.~Usunier, A.~Kirillov, and S.~Zagoruyko,
``End-to-End Object Detection with Transformers,''
in \textit{Proc. Eur. Conf. Comput. Vis. (ECCV)}, 2020.

\bibitem{Schmidt2025}
S. Schmidt \textit{et al.}, ``Prior2Former - Evidential Modeling of Mask Transformers for Assumption-Free Open-World Panoptic Segmentation,'' in \textit{Proc. IEEE/CVF Int. Conf. Comput. Vis. (ICCV)}, 2025.


\end{thebibliography}
\end{document}